# Predictors of disease outbreaks at continental-scale in the African region: Insights and predictions with geospatial artificial intelligence using earth observations and routine disease surveillance data

**Scott Pezanowski[1], Etien Luc Koua[2], Joseph C Okeibunor[2] and Abdou Salam Gueye[2]**

## Abstract

**Objectives:** Our research adopts computational techniques to analyze disease outbreaks weekly over a large geographic area while maintaining local-level analysis by incorporating relevant high-spatial resolution cultural and environmental datasets. The abundance of data about disease outbreaks gives scientists an excellent opportunity to uncover patterns in disease spread and make future predictions. However, data over a sizeable geographic area quickly outpace human cognition. Our study area covers a significant portion of the African continent (about 17,885,000 $km^2$). The data size makes computational analysis vital to assist human decision-makers.

**Methods:** We first applied global and local spatial autocorrelation for malaria, cholera, meningitis, and yellow fever case counts. We then used machine learning to predict the weekly presence of these diseases in the second-level administrative district. Lastly, we used machine learning feature importance methods on the variables that affect spread.

**Results:** Our spatial autocorrelation results show that geographic nearness is critical but varies in effect and space. Moreover, we identified many interesting hot and cold spots and spatial outliers. The machine learning model infers a binary class of cases or none with the best $F1$ score of 0.96 for malaria. Machine learning feature importance uncovered critical cultural and environmental factors affecting outbreaks and variations between diseases.

**Conclusions:** Our study shows that data analytics and machine learning are vital to understanding and monitoring disease outbreaks locally across vast areas. The speed at which these methods produce insights can be critical during epidemics and emergencies.

## Introduction

Computational methods like machine learning are becoming vital to analyzing disease spread because they are excellent tools for quickly finding patterns in large and complicated datasets.[1,2] Our study investigated geospatial technologies, artificial intelligence (AI), and data analytics

[1]BrightWorld Labs, State College, PA, USA
[2]Emergency Preparedness and Response, WHO Regional Office for Africa, Brazzaville, Congo

**Corresponding author:**
Scott Pezanowski, BrightWorld Labs, 618 Wayland Pl., State College, PA 16803, USA.
Email: scottpez@brightworldlabs.com

methods to analyze disease outbreaks and their contributing factors throughout the vast jurisdiction of the World Health Organization (WHO) Regional Office for Africa (WHO AFRO).

The land area covered by WHO AFRO is approximately 17,885,000 $km^2$. To put this into perspective, the size of the contiguous United States of America (without Alaska and Hawaii) is 8,080,464 $km^2$, or 45% of the WHO African Region. Furthermore, the size of New York State is about 141,300 $km^2$ or 0.79% of the area. Moreover, Central Park is about 3.41 $km^2$. Visualizing all of Central Park while standing in it is challenging. Now, imagine 5,244,868 Central Parks. Throughout this massive land area, one can only begin to comprehend the diversity of the approximately 1,124,700,000 people, transportation networks, schools, hospitals, landforms, water bodies, vegetation, weather, and climate events. Finally, let us visualize that most people, places, things, and events regularly change over time.

Visualizing these vast areas and differing characteristics can give us a sense of the challenge that WHO AFRO faces in monitoring, analyzing, understanding, preparing for, and predicting disease outbreaks throughout the Region. Despite its many successes in mitigating disease outbreaks, WHO AFRO will benefit from modern AI and data analytics that can uncover patterns in disease outbreaks, derive critical attributes affecting outbreaks, predict and model future outbreaks, and detect and monitor outbreaks through media.

Since our study area is massive, our analysis and predictions are weekly and at the local level, and we include many diseases, we must surpass previous research to analyze and monitor disease outbreaks. When researchers analyze data at a fine geographic granularity (like districts), one method they use to circumvent computational method deficiencies is either limiting the study area to a relatively small area or limiting the analysis to one disease. Starting with studies to use machine learning to predict disease, Atek et al.[3] predicted spatial patterns of the West Nile virus in Italy using machine learning while grouping data by a few sub-country regions. Kim et al.[4] forecasted multiple diseases in South Korea. Husin et al.[5] used a machine learning neural network to predict dengue in Malaysia, while Usmani et al.[6] predicted cholera cases in Yemen. Bellocchio et al.[7] forecasted COVID-19 in Europe, and Tian et al.[8] tested multiple machine learning models to predict meningitis in Nigeria. Wang et al.[9] predicted spatial and temporal trends in infectious diseases in China. Also, in China, Xu et al.[10] took a slightly different approach by using long short-term memory (LSTM) time-series predictions for dengue fever based on the temporal patterns of past cases. These valuable studies using machine learning to predict disease spread focused on individual countries.

Some literature analyzed spatial correlations and distributions, but, again, only across relatively small geographic areas. Sirisena et al.[11] used spatial autocorrelation to find spatial relationships of dengue in Sri Lanka. Ali et al.[12] identified spatial patterns and risk factors for cholera in Bangladesh.

When data covers a sizeable geographic area or even the entire globe, past studies aggregated the data into large divisions like countries. Dixon et al.[13] compared machine learning forecasting methods of infectious disease using country-level data for a few countries globally. Also, Liang et al.[14] predicted global African swine fever outbreaks at the country level using machine learning. Finally, Hess et al.[15] provided a thorough global spatial analysis of West Nile virus.

As data has grown, computational techniques have improved that can surpass the literature by maintaining fine geographic scales in analyses and predictions over large areas across many diseases. Our research aims to explore computational methods that have the advantage of performing continental-scale analyses while maintaining detailed geographic granularity, as advocated for by Koua and Kraak,[16] Desai et al.,[17] Gwenzi and Sanganyado,[18] and Neill.[19] Moreover, once developed and created, our analyses can be re-run in fractions of a second. Obtaining our goals means that international organizations and governments will have enhanced tools to understand, prevent, address, and recover from outbreaks.

We organize our article by the current Introduction section followed by the Methodology section. We begin the Methodology section by discussing our data sources. For disease data analysis, our primary data is the WHO AFRO Integrated Disease Surveillance and Response (IDSR) dataset which provides our independent variable of various disease suspected cases and one dependent variable, the week of outbreaks. Next, we report on the dependent variables that represent factors we believe influence outbreaks and processing steps to derive them from multiple high-spatial-resolution datasets.

After describing our data, we detail our computational analysis methods, beginning with exploratory spatial data analysis (ESDA) and global spatial autocorrelation to understand the importance of nearness. We also used local spatial autocorrelation to uncover detailed relationships and patterns like hot and cold spots. Our second method involved predicting disease spread using machine learning. Our third method employed machine learning feature importance techniques to derive the vital contributing factors to outbreaks.

We highlight our key research findings in our Results section. First, geographic proximity strongly influences disease outbreaks but at varying levels per disease. Second, individual diseases show different geographic dispersion patterns with distinctive local hot and cold spots and outliers. We identified multiple local geographic locations that offer fascinating patterns. Above and beyond the direct ESDA findings, the results dictated some of the contributing datasets we acquired for our machine learning methods. Third, our machine learning model trained on our dependent datasets successfully predicts our independent, district-level disease cases at various rates for four diseases on a weekly basis. Our best result predicted the weekly presence of malaria with an $F1$ score of 0.96. Fourth, we found that the dependent datasets of disease-contributing factors were of differing importance in predicting these four diseases.

We conclude our article with a Discussion section where we interpret our geospatial and machine learning findings and suggest future work and datasets to add to our analysis. One proposed future study entails performing a detailed local analysis using more detailed supplementary datasets like Zheng et al.[20] and Li et al.,[21] which can be scaled up to a larger geographic area.

Although some of our findings about disease patterns are not unique, our novel application of geospatial analysis and AI to comprehensive outbreak data combined with high spatial and temporal-resolution continental-scale data predictors can significantly contribute to data-driven global public health monitoring and outbreak control, prevention, and response. Furthermore, we must emphasize again that our research contributions go beyond existing literature[23,22] because our computational methods are scalable geographically and temporally to large areas while maintaining local-level analysis across many diseases. Moreover, they can be performed quickly, allowing for rapid results that can be repeated regularly, frequently, and in crisis events.

## Methodology

Our study investigated how modern computing techniques can aid the understanding, monitoring, and prediction of disease outbreaks. In this section, we first describe the datasets and respective processing steps we used for our study, divided into a Disease surveillance data and selection subsection about disease case counts and a Sources and processing of datasets that affect outbreaks subsection where we list datasets that can impact the spread of disease. Then, we describe our initial methods to uncover spatial patterns in the ESDA for spatial patterns and relationships subsection. Next, we employed data science and machine learning computational techniques to predict the presence of disease based on these case counts from environmental and demographic data described in the Machine learning methods to predict disease cases subsection. Lastly, the Machine learning methods for feature importance subsection describes machine learning techniques to identify the most critical contributing factors to predicting outbreaks.

We collected the disease surveillance data from the third edition of the IDSR dataset, which was gathered and curated between January 2019 and December 2022. We processed, analyzed, and interpreted all the data between June 2022 and October 2023. All research for this study was based out of WHO AFRO in Brazzaville, Congo.

### Data sources and processing steps

The WHO AFRO IDSR dataset is our primary dataset that tracks infectious diseases. To support our macro geographic scale analysis, we added many additional high-spatial-resolution datasets to help us understand the spatial and temporal patterns of disease outbreaks. In statistical terms, the IDSR dataset case counts are the dependent variable we analyze and try to predict with machine learning, while time and the supplementary datasets are independent variables.

*Disease surveillance data and selection.* The WHO AFRO IDSR dataset records suspected cases and deaths from various diseases at the second administrative district level on a weekly basis throughout the WHO African Region. The dataset covers all countries under the jurisdiction of WHO AFRO (that is most of the African continent). At the time of our research, Algeria, Angola, Mauritius, and South Africa did not participate in IDSR. Figure 1 shows the countries monitored with their population density in green (darker shades mean denser) and bar charts for each country with log-normalized total deaths and cases in purple and brown, respectively. The figure summarizes the variability in disease spread throughout the Region and outlines the spatial extent. We did not include a map legend since we normalized the case and death counts to show relative amounts. The Democratic Republic of the Congo has the highest number of total disease cases (109,384,201), whereas Lesotho has the lowest number with 825. The mean number of cases per country is 12,397,797, while the first quartile, median, and third quartile case counts are 76,637, 693,795, and 13,583,939, respectively. Therefore, cases aggregated by country are right-skewed, where a few large values significantly raise the mean. Our initial analysis focused on cases and not deaths since cases provide a good indication of disease distribution.

The IDSR dataset we used was exported at the end of 2022 and contained entries from 1 January 2019 through 24 December 2022 (209 weekly entries). Furthermore, there are entries for every second-level administrative district (4506 districts). Moreover, IDSR records most diseases present in the study area. Table 1 shows a selected sample of five rows to illustrate the dataset. Each row corresponds to the number of cases and deaths per disease occurring in each administrative district in each week of the year. Thus, if two diseases occur in two districts over 2 consecutive weeks, this will produce eight rows, as shown in equation (1).

$$2\ diseases \times 2\ districts \times 2\ weeks = 8\ rows \tag{1}$$

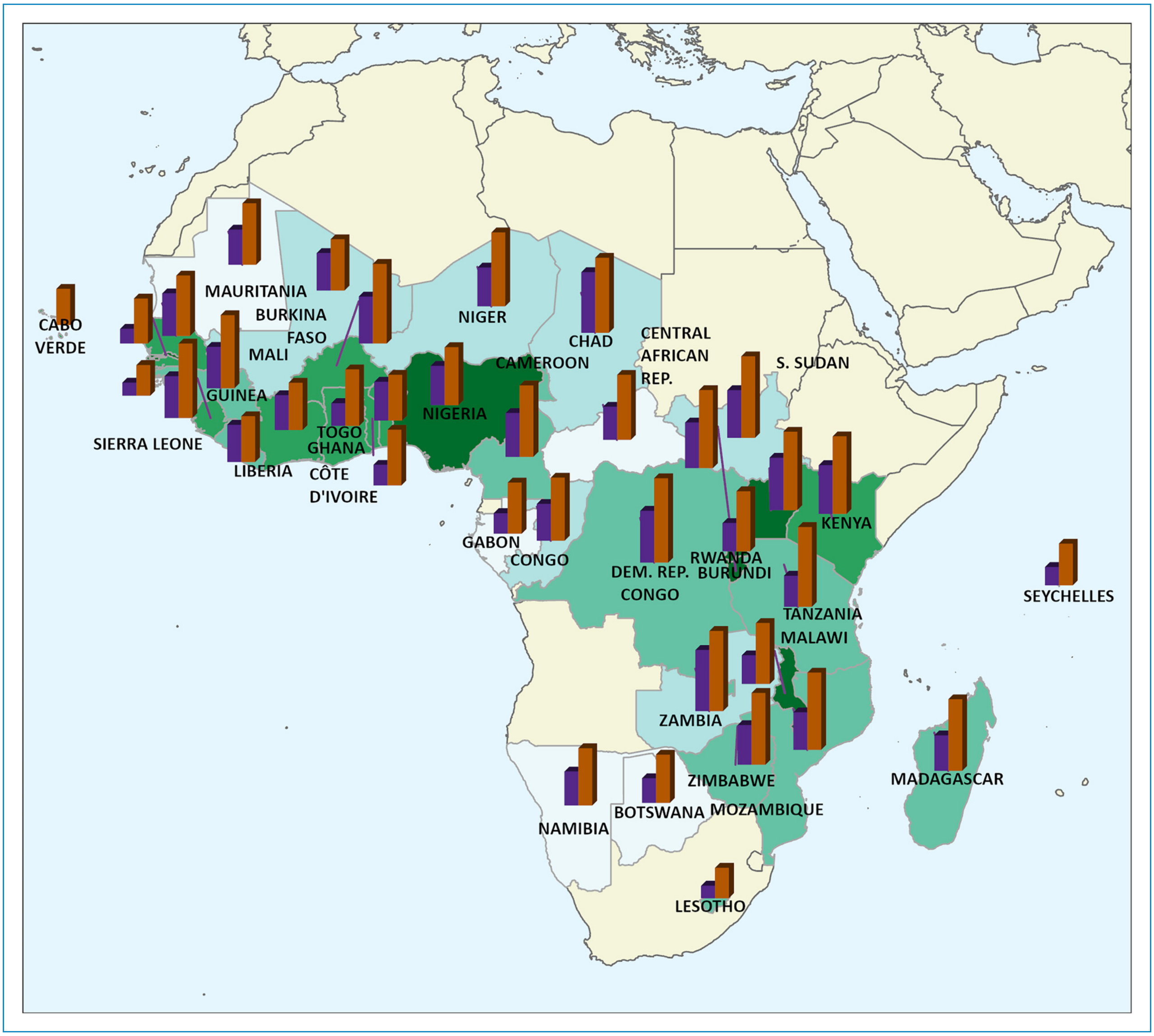

**Figure 1.** Countries monitored by WHO AFRO with their population density in green (darker shades mean denser) and bar charts with normalized total deaths and cases derived from the IDSR dataset in purple and brown bars, respectively. WHO AFRO: World Health Organization Regional Office for Africa; IDSR: Integrated Disease Surveillance and Response.

We chose the four diseases (malaria, cholera, meningitis, and yellow fever) because they are prevalent and severe on the continent. They are also infectious diseases and are known to be influenced by many of the environmental factors we chose to focus on in our analysis. When we queried the IDSR dataset for geospatial analysis and machine learning, we queried cases for only one disease at a time.

*Sources and processing of datasets that affect outbreaks.* The supporting datasets used to analyze disease outbreaks with machine learning include several continent-wide, high-resolution datasets. These datasets include elevation, land cover, precipitation, temperature, population density, and relative wealth estimates. Other research projects similarly attempted to use environmental and demographic data to predict diseases, but each focused only on small geographic areas or single diseases.[12,7,18,15,4,14,11,6,20,24]

We chose our dependent variables because of their known influence on many infectious disease outbreaks. Furthermore, because climatic factors are changing, we would like a way to model the potential impact of climate change on disease outbreaks. Moreover, our local indicators of spatial association (LISA) results (presented in Figure 2 and the "Spatial proximity's impact on disease outbreaks" subsection) showed a strong impact of climate on disease outbreaks. Our goal in this study was not to uncover new factors influencing disease outbreaks. Instead, methods

**Table 1.** A five-row sample of the Integrated Disease Surveillance and Response (IDSR) dataset for reference.

| Year | Week | Country | Province | District | Disease | Number of cases | Number of deaths |
|---|---|---|---|---|---|---|---|
| 2019 | 1 | Burundi | Bururi | Matana | Malaria | 511 | 1 |
| 2019 | 2 | Benin | Atacora | Cobly | Rabies | 0 | 0 |
| 2019 | 14 | Côte d'Ivoire | Worodougou | Seguela | Measles | 1 | 0 |
| 2019 | 43 | Democratic Republic of the Congo | Nord-Kivu | Oicha | Typhoid fever | 59 | 0 |
| 2019 | 43 | Democratic Republic of the Congo | Nord-Kivu | Kibirizi | Severe acute respiratory infections (SARIs) | 896 | 0 |

must be developed to analyze and monitor their influence on disease over the vast study area.

Some dependent variables were considered and tested but ultimately excluded because of inadequate spatial or temporal resolution. For example, water, sanitation, and hygiene (WASH) data shows sanitary measures but only aggregates to the country level.[25] Using WASH data in machine learning to predict disease at the district level would not be appropriate.

In summary, data was included and excluded because of its potential effect on the disease, its current importance in our environment, and its availability, quality, and temporal and spatial fit with our analysis. The supporting datasets are summarized in Table 2.

After choosing and gathering the supporting datasets, we needed to perform processing to transform them into data suitable for our machine learning challenge. We extracted multiple land cover classifications from the single land cover dataset. For each land cover class extracted, we calculated the total area covered and the percentage of the district covered. Then, for each class, we used a min–max scaler to scale the area covered, percent covered, and relative population. We then proceeded to add the three scaled values together to produce the land cover class value used in machine learning. We calculated this to ensure that large districts with a small percentage of area covered, small districts with a large percentage of area covered, or districts with a large land cover in a class where people do not live were assigned the correct weight by the model. The land cover classes we extracted include:

- vegetation
- crops
- built-up areas
- bare ground
- and rangeland

As our final feature, in addition to overall population densities, we calculated each district's population close to water bodies. We added this feature because many infectious diseases have a higher transmission rate when they occur near water bodies (either directly through the water itself or because the water promotes insects that transmit the disease). To compute the population for each district near a water body, we downloaded OpenStreetMap (OSM) data for Africa and selected water bodies (using the OSM tags "natural" = "water" OR "water" IS NOT NULL). This query produced vector features for all inland water bodies and rivers. Next, we computed 1, 3, and 6 km buffers around every inland water body using the PostGIS ST_Buffer function. We chose multiple distances for the test since we did not know precisely the extent of the influence of the water body. Then, we converted the vector buffers to a raster that matched the population dataset in resolution. We used the ArcGIS Pro Conditional tool[33] to create a new raster with the population density where the buffer overlaps with a zero value outside the buffer.

We acquired all supporting datasets as continuous rasters, except for relative wealth, because this is distributed as point data where points correspond to a significant population. To match the spatial resolution of these datasets with the IDSR districts, we aggregated them with the mean value within each district to produce a single value. To aggregate them, we used the ArcGIS Pro Zonal Statistics as Table tool,[34] which calculates single summary values for each district per dataset. Because the land cover dataset is multiclass, we used the ArcGIS Pro Tabulate Area tool,[35] which produces the same result but on multiclass data.

Finally, we joined the IDSR dataset using the week and district. For datasets updated less than weekly, we joined them in a one-to-many relationship. For example, the single land cover value per year was assigned to each data record in that corresponding year. In contrast, we computed an aggregate value for datasets updated more frequently than a week (such as the mean temperature). Table 3shows a sample of the data after these processing steps (described in the "Machine learning methods to predict disease cases" subsection).

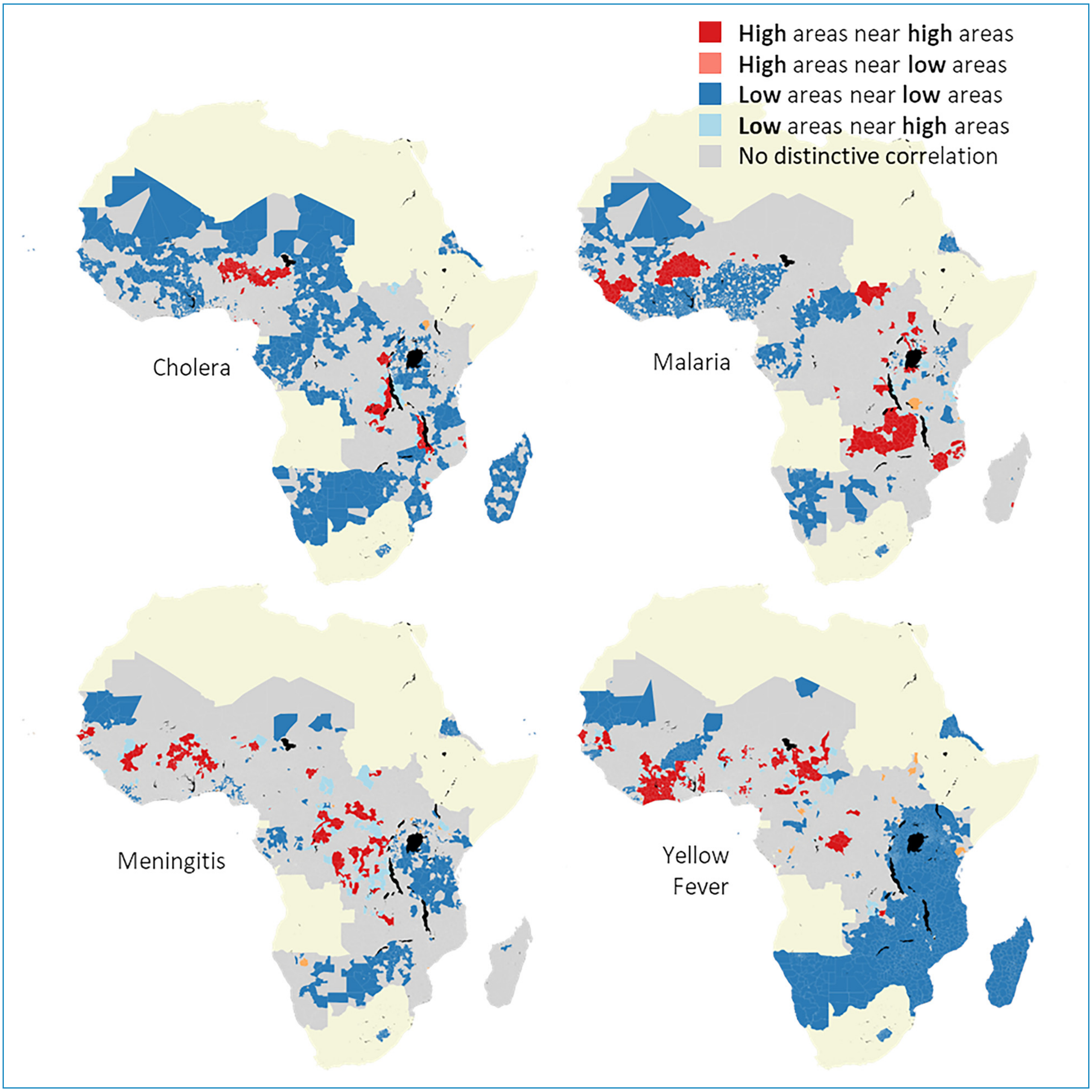

**Figure 2.** Moran's local analysis shows hot spots in dark red, cold spots in dark blue, low cases around hot spots in light blue, and high cases around low areas in light red.

## *ESDA for spatial patterns and relationships*

One of the aims of our research was to find patterns in disease spread to better understand affected areas and the factors that influence disease spread over large geographic areas. Accordingly, our first step was to perform ESDA. The primary purpose of exploratory data analysis is to summarize and analyze the data without assumptions. To this end, we primarily used unsupervised machine learning techniques where computer models look for similarities and differences in the data without human intervention. Our ESDA closely followed tutorials from the valuable online class "A Course on Geographic Data Science" by Arribas-Bel.[36]

To group data records and find spatial patterns, we used spatial autocorrelation. Spatial autocorrelation is a technique that relies on the existence of a "functional relationship between what happens at one point in space and what happens elsewhere."[37] Events occurring in one location are considered in addition to spatial weights, meaning that nearby values are more vital than farther values. If the actual pattern of values follows this trend, we can conclude that the values are spatially autocorrelated. Spatial autocorrelation has broad applications, including some infectious disease-related analyses.[38]

**Table 2.** Supporting datasets used in machine learning to make disease case count predictions.

| Data attribute | Name | Provider | Frequency |
|---|---|---|---|
| Elevation | Advanced Spaceborne Thermal Emission and Reflection Radiometer (ASTER)[26] | NASA Jet Propulsion Laboratory (JPL) | One time (2000–2013) |
| Population density | Unconstrained individual countries 2000–2020 (1 km resolution)[27,28,24] | WorldPop | Two times (2020 & 2022) |
| Relative wealth | Relative wealth index[29] | Meta | One time (2022) |
| Land cover | 10-m Annual Land Use Land Cover (9-class)[30] | Microsoft | yearly |
| Precipitation | GPM IMERG Early Precipitation L3 one day $0.1° \times 0.1°$ V06[31] | NASA | Two times per day |
| Temperature | GLDAS Noah Land Surface Model L4 monthly $0.25 \times 0.25°$ V2.1[32] | NASA | Two times per day |

To prepare the data for the spatial autocorrelation algorithm, we grouped all IDSR records by district and disease and summed the cases. The disease cases are the only column considered for spatial autocorrelation. For example, a table for cholera has 4019 rows, representing the number of districts in the study area. Each row only contains the total number of cholera cases for that district over the 5 years of the IDSR dataset and the spatial geometry of the boundary of that district. We read this data table from the database into the GeoPandas Python library.[39]

First, we calculated the global spatial autocorrelation for each disease using Moran's $I$.[40] Moran's $I$ produces a value that measures the overall degree of spatial autocorrelation. Scientists typically interpret Moran's $I$ analysis from an $I$ value showing the global spatial autocorrelation and a statistical $p$-value indicating the significance of the spatial autocorrelation compared to random chance.[41] The $I$ value ranges between $-1$ and 1. "A value of $-1$ is perfect clustering of dissimilar values, 0 is no correlation, and $+1$ indicates perfect clustering of similar values."[42]

Furthermore, we wanted to analyze local differences in spatial autocorrelation's because the vast study area and a global spatial autocorrelation do not show the entire picture. Therefore, we used Moran's local functionality of the ESDA library to compute LISA.[43] This calculation produces a value for every administrative district. The value indicates whether the district falls into one of four categories compared with its bordering neighbors. The four categories are:

- HH—high values surrounded by high values
- LL—low values surrounded by low values
- HL—high values among low values
- LH—and low values among high values

HH can be considered spatial hot spots and LL low spots, whereas the latter two categories are spatial outliers.

## Machine learning methods to predict disease cases

After our ESDA, we investigated the potential for a machine learning model that can predict weekly cases of diseases by district. To do this, we grouped the IDSR dataset differently from the spatial autocorrelation analysis. Here, we filtered the records by the same four individual diseases but left the weekly district records as they were. After filtering the records for a disease, we joined the supporting datasets with the district geographic dataset. Table 3 shows a sample of five rows of the data used for machine learning (the first eight columns at the top and the remaining 10 columns below). The ID column is a unique identifier for each row, whereas the ADM_ID column corresponds to an internal identifier for the administrative district geographic dataset. There are 17 features in each row and one column showing disease cases.

We previously mentioned that we calculated the population near water bodies at three distances. We tested the three distances using our machine learning feature importance measures (described in the "Uncovering disease contributing factors with feature importance" subsection) to decide which to use. The feature importance plot showed that the three population-near-water columns were primarily redundant, with the 3 km buffer being slightly more important consistently. Therefore, only populations within 3 km of an inland water body were used to train the model.

Before training the model on this data, we combined the two columns for each land cover into one (as described in the "Sources and processing of datasets that affect outbreaks" subsection). As a result, the ten columns for land cover became five, and after processing, a model trained on the 12 independent variables attempted to predict the target disease cases. We scaled our data using the scikit-learn Python library Robust Scaler because the data has outliers.

**Table 3.** A five-row sample of data used for machine learning, with the first eight columns at the top and the next 10 columns below. The land cover area values are the number of raster cells in the district within that land cover class.

| ID | ADM_ID | Week | Precipitation | Temperature | Trees (area) | Trees (%) | Crops (area) | Crops (%) | Built-up (area) |
|---|---|---|---|---|---|---|---|---|---|
| 2157 | 6513 | 1/1/2019 | 0.02 | 31.68 | 0 | 0 | 13 | 0.15663 | 67 |
| 2160 | 5763 | 1/1/2019 | 8.56 | 29.09 | 11,215 | 0.79 | 5 | 0.00035 | 25 |
| 2161 | 5788 | 1/1/2019 | 0 | 35.17 | 10,263 | 0.88 | 1 | 0.00009 | 28 |
| 2162 | 1079 | 1/1/2019 | 16.28 | 28.13 | 17 | 0.05 | 6 | 0.01929 | 246 |
| 2163 | 6322 | 1/1/2019 | 0.52 | 30.01 | 144 | 0.65 | 1 | 0.00448 | 69 |

| ID | Built-up (%) | Bare ground (area) | Bare ground (%) | Rangeland (area) | Rangeland (%) | Relative population density | Relative population near water | Relative wealth | Elevation | Total cases |
|---|---|---|---|---|---|---|---|---|---|---|
| 2157 | 0.807 | 1 | 0.012 | 1 | 0.01 | 6653 | 0 | 0 | 18 | 1 |
| 2160 | 0.002 | 0 | 0 | 2691 | 0.19 | 54 | 156,404 | 0.9 | 617 | 1 |
| 2161 | 0.002 | 0 | 0 | 1150 | 0.10 | 18 | 115,419 | 0.8 | 522 | 1 |
| 2162 | 0.791 | 4 | 0.013 | 17 | 0.06 | 1317 | 6973 | 2.5 | 23 | 2 |
| 2163 | 0.309 | 1 | 0.004 | 8 | 0.04 | 379 | 66,592 | 1.7 | 693 | 1 |

Since treating this machine learning problem as regression and attempting to predict the exact case counts would be challenging, we converted the case counts to a binary classification of cases versus no cases in this initial research. This binary classification is still a substantial first effort, and future regression machine learning work can be done with more comprehensive supporting datasets and improved IDSR data.

In summary, the time range covers 209 weeks for 4506 districts. Each disease was queried separately, producing four tabular datasets with 12 independent features and one dependent binary variable for every week and district. Moreover, each machine learning model was trained on 209 weeks $\times$ 4506 districts $=$ 941,754 records.

Throughout the vast study area and time range, it was much more common for a district to have no disease than to have cases. As a result, each training dataset is imbalanced. Table 4 summarizes the imbalanced data for the four diseases, spotlighting the malaria dataset as the most balanced, with 32.3% positive records, and the dataset for cholera as the most imbalanced, with 1.6% positive records.

We trained the machine learning model on a random 80/20 data split, where we held 20% of the records for testing. We used the AutoGluon Automated Machine Learning (AutoML) library[44] and its TabularPredictor. AutoML libraries use automated techniques to process data, pick the best-performing machine learning model, and optimize the machine learning model's hyperparameters.

We trained the model to account for class imbalance and used the $F1$ score as the evaluation metric. We tried random undersampling and oversampling with SMOTE[37] to handle the class imbalance. However, training the model with all data and the $F1$ score as the evaluation metric produced the best results.

## *Machine learning methods for feature importance*

In addition to machine learning predictions, we used machine learning feature importance measures to identify features that are relatively valuable and not valuable in predicting disease cases. Feature importance has been used to determine real-world influencers in the spread of disease, as they were by Xiong et al.[45] and Mihoub et al.[46] to understand the influencers of COVID-19. AutoGluon uses a permutation importance method to calculate feature importance.[47] In brief, a permutation importance score is produced when the model is trained and makes predictions using perturbed or randomly shuffled dataset rows.

**Table 4.** Summary of positive and negative records for each disease illustrating the class imbalance challenge.

| Disease | No. of records with cases (positive) | No. of records with no cases (negative) | Total records | Percent positive class |
|---|---|---|---|---|
| Cholera | 14,676 | 927,078 | 941,754 | 1.6% |
| Malaria | 304,467 | 637,287 | 941,754 | 32.3% |
| Meningitis | 30,957 | 910,797 | 941,754 | 3.3% |
| Yellow fever | 27,595 | 914,159 | 941,754 | 2.9% |

## Results

This section describes our results in the subsections on ESDA spatial autocorrelation, machine learning modeling to predict disease case counts, and machine learning feature importance measures.

### *Spatial proximity's impact on disease outbreaks*

Once the data was transformed into a clean table for analysis, we used the ESDA Python library to run a Moran's $I$ analysis for global spatial autocorrelation on the four diseases. Table 5 summarizes the results of these analyses. ESDA's Moran's $I$ analysis was valuable in determining that all four diseases showed a statistically significant positive global spatial autocorrelation. Malaria had the strongest spatial autocorrelation with an $I$ value of 0.2442. A straightforward interpretation of the results from applying Moran's $I$ on our data is that districts with high case counts are likelier to be near other districts with high case counts, and vice versa.

Since Moran's $I$ only produces a global spatial autocorrelation value, we ran LISA, which resulted in a spatial autocorrelation value for every district across our large study area. We visually analyzed the results on the local district maps in Figure 2. These maps show the spatial patterns of the four diseases across the study area, with striking hot and low spots and outliers that vary between the four diseases. Hot spots are in dark red, cold spots are in dark blue, low cases around hot spots are in light blue, and high cases around low areas are in light red. The latter two categories are spatial outliers.

What we found most striking about these maps are the different patterns between the four diseases. Cholera has two relatively small hot spots in the Democratic Republic of the Congo and Nigeria, with many large cold spots. Malaria has more hot spots that are distinctively separate from the cold spots. Meningitis has many geographically small hot spots, a pronounced hot spot in Côte d'Ivoire, and some outliers scattered throughout the maps. Yellow fever has many small hot spots, primarily in the savanna areas of Africa, and a large cold area covering a mostly continuous area in the south.

**Table 5.** Summary of Moran's $I$ analysis for global spatial autocorrelation of four diseases. Cholera and malaria show strong positive spatial autocorrelation, whereas yellow fever is lower but still positive and significant.

| Disease | $I$ | $p$ |
|---|---|---|
| Cholera | 0.2434 | 0.001 |
| Malaria | **0.2442** | 0.001 |
| Meningitis | 0.1140 | 0.001 |
| Yellow fever | 0.0895 | 0.001 |

Our local spatial autocorrelation results produced distinctive maps showing that climate is vital to outbreak predictions. Two examples of climate importance from Figure 2 include malaria hot spots in tropical climates and yellow fever hot spots in savanna climates. Therefore, our initial spatial epidemiological analysis led us to include precipitation, temperature, and elevation in our machine learning analysis.

The LISA map for cholera also showed a fascinating area in the Democratic Republic of the Congo, namely in the Congo River basin extending east into Burundi, Rwanda, Uganda, and Tanzania. Furthermore, this area has a pronounced hot spot to the west of Lake Tanganyika, which is considered the headwaters of the Congo River. Nevertheless, there are also distinctive cold outliers within this hot spot. Moreover, to the east of the Great Lakes are definitive cold spots. Thus, in our Proposed large geographic scale analysis section below, we recommend detailed analyses in the future for this area.

### *Predicting weekly disease cases by district*

Our predictive model for the binary class of district-level weekly disease cases for malaria was the most successful,

with an $F1$ score of 0.96. Based on our experience, in addition to the chosen features being good predictors of malaria, the high $F1$ score is also explained by the fact that malaria had the most significant number of positive records (least imbalanced), allowing the model to learn more. Cholera had an $F1$ score of 0.65. The model predicted a binary class for meningitis with an $F1$ score of 0.41, whereas the metric for yellow fever was lower at 0.17. Unlike malaria, which showed many positive cases, we conjecture that the performance of the other disease models suffered from the lack of positive records. With less training data than malaria, the models were less effective in learning the data patterns and inferring cases.

Table 6 summarizes the evaluation metrics used to classify the best-performing machine learning models for the four diseases. All top models had a very high level of accuracy. However, accuracy is not a solid metric in our case because the data is imbalanced, and although the model predicts the no-case class well, it often falls short in predicting the case class. Since predicting districts with cases is essential, we must focus on the $F1$ score to ensure both classes are accurately inferred.

Table 7 shows the top six performing machine learning models, as reported by AutoGluon's leaderboard function. First, the metrics reported in Table 6 were derived from the top-performing models. Second, these lists provide information on which model types are best suited for the data. Interestingly, AutoGluon preferred non-deep learning models over deep learning models. We hypothesize that this is because of the way we set up the machine learning problems, and with the class imbalance, there was insufficient training data for deep learning. Cholera had fewer positive records, and AutoGluon preferred simpler models like random forest. Meanwhile, the best models for malaria are more complex.

### *Uncovering disease contributing factors with feature importance*

As already mentioned, we used AutoGluon's feature importance function. The function hints at how to improve the model predictions and provides real-world clues as to what factors are the most critical influencers of disease spread. These feature importance measures can provide evidence to support better decisions about future outbreak response and prevention efforts. Figure 3 shows the model's feature importance measures for each disease.

A comparison of the model's feature importance measures of the four diseases shows time as a significant factor in disease prediction. For cholera, malaria, and meningitis, tree coverage is essential. We can see that elevation is crucial for each disease; however, this result requires more analysis to confirm it because it is the only dataset with a single value that we repeated across time. A consistent value like this could cause issues with the machine learning model. Nonetheless, the result does suggest the importance of elevation since it will not change much in a few years. It is important to note that the numerical values along the horizontal axis of the charts are relative. Therefore, these values have no significance when comparing the four disease charts.

## Discussion

Our research towards initial massive geospatial analytics and machine learning applications to disease spread in Africa showed four critical points for discussion. First, Tobler's First Law of Geography applies to disease spread since there is a significant spatial autocorrelation between districts in the four diseases. Tobler's First Law states, "Everything is related to everything else, but near things are more related than distant things."[48] Therefore, we can apply this law to our results, showing that districts near other districts where the disease is present are likelier to have that disease and vice versa. However, Tobler's First Law is clear that spatial proximity is not the only contributing factor. Our study area's diverse local hot and cold spots, which we uncovered with local spatial autocorrelation, clearly illustrate this point.

After spatial analysis, our second discussion topic is that a machine learning model can accurately predict the presence or absence of district-level weekly disease cases for multiple diseases. However, some diseases were not as successful, and this can be because the chosen features are not

**Table 6.** Model evaluation metrics for machine learning to predict the weekly presence of all four diseases by district.

| Disease | Accuracy | Balanced accuracy | MCC | ROC AUC | $F1$ | Precision | Recall |
|---|---|---|---|---|---|---|---|
| Cholera | **0.991** | 0.772 | 0.652 | 0.971 | 0.646 | 0.788 | 0.546 |
| Malaria | 0.973 | **0.975** | **0.939** | **0.995** | **0.959** | **0.938** | **0.981** |
| Meningitis | 0.971 | 0.651 | 0.431 | 0.931 | 0.416 | 0.636 | 0.309 |
| Yellow fever | 0.971 | 0.548 | 0.224 | 0.897 | 0.167 | 0.553 | 0.098 |

MCC: Matthews correlation coefficient; ROC AUC: receiver-operating characteristic curve area under the curve.

**Table 7.** The top six performing machine learning models for each disease as reported by AutoGluon's leaderboard. We used the top model for each disease to make our predictions and report evaluation metrics.

| Cholera | Malaria | Meningitis | Yellow fever |
|---|---|---|---|
| 1. RandomForestEntr | 1. LightGBM | 1. RandomForestEntr | 1. RandomForestGini |
| 2. RandomForestGini | 2. LightGBMLarge | 2. RandomForestGini | 2. WeightedEnsemble_L2 |
| 3. CatBoost | 3. WeightedEnsemble_L2 | 3. WeightedEnsemble_L2 | 3. CatBoost |
| 4. WeightedEnsemble_L2 | 4. XGBoost | 4. NeuralNetTorch | 4. NeuralNetTorch |
| 5. NeuralNetTorch | 5. RandomForestEntr | 5. CatBoost | 5. XGBoost |
| 6. LightGBMXT | 6. RandomForestGini | 6. ExtraTreesEntr | 6. ExtraTreesEntr |

comprehensive enough to predict the disease or because there are not enough positive records. Furthermore, these initial results are promising in terms of future analyses since they occurred, even though we only used a few supporting datasets in our study.

Third, a critical discussion point is that a machine learning model can produce feature importance measures to determine the most essential factors in predicting the spread of disease.

The fourth key point is that although traditional methods could be used in some of our analyses, modern methods have clear advantages. Modern techniques can scale analysis over vast geographic areas while maintaining local-level knowledge. Furthermore, modern techniques can produce results in milliseconds, while previous methods might take months or much longer.

In addition to these four discussion points, we emphasize an overarching research finding that space and time are vital in analyzing, predicting, and monitoring disease spread. Our extensive contributions from our initial study can significantly impact future analysis, and in the subsections below, we identify essential steps to make this possible. Nonetheless, we must discuss our study's limitations next.

The primary limitation of this initial study is that we focused on characterizing and predicting outbreaks of the four chosen diseases as measured in the IDSR dataset. Although IDSR is comprehensive and extensive, it presents summarized case counts and does not include detailed information that might help determine why outbreaks, patterns, or outliers occur. Even though our study is limited to this data, based on our experience, we surmise that our techniques would be effective for similar data with similar spatial and temporal resolutions.

A secondary limitation of our study is that because we focused on developing extensible large-scale analysis and monitoring techniques, we limited our acquisition of supporting data to factors specific to the chosen diseases. Therefore, an additional limitation is that our analysis only applies to the four diseases. However, based on our results, we are confident that our techniques will be effective in analyzing and predicting other diseases once more relevant data is incorporated for those diseases.

## *Insights to improve predictions of disease outbreaks*

To provide a more detailed discussion, first, we draw attention to the machine learning model for predicting malaria, which performed significantly better than the other three models. We must note this because it means that our chosen data features are good predictors of malaria. Also, we surmise that the very high $F1$ score is because malaria had the highest number of positive cases, allowing the model to learn the data patterns. These results strengthen the importance of improved comprehensiveness and accuracy of the IDSR dataset over time. In addition, the model's weak performance for diseases with fewer case counts in the IDSR dataset again highlights that improvements in IDSR data collection and added data over time will improve any analytics performed with it.

Along with highlighting variations in model performances, it is critical to emphasize that time is consistently crucial in predicting cases. The importance of time to our model suggests that future machine learning to predict disease would benefit from a greater focus on time. One way would be to treat the IDSR dataset as time series data and then use machine learning techniques and models ideal for time series data.

Despite our promising machine learning results, some mixed results mean more data processing, gathering relevant supporting datasets, and incorporating modern deep learning models will likely improve outcomes significantly. It is critical in future work to add more features (relevant datasets) and ensure the analysis is repeatable. With more future data and greater data accuracy, future

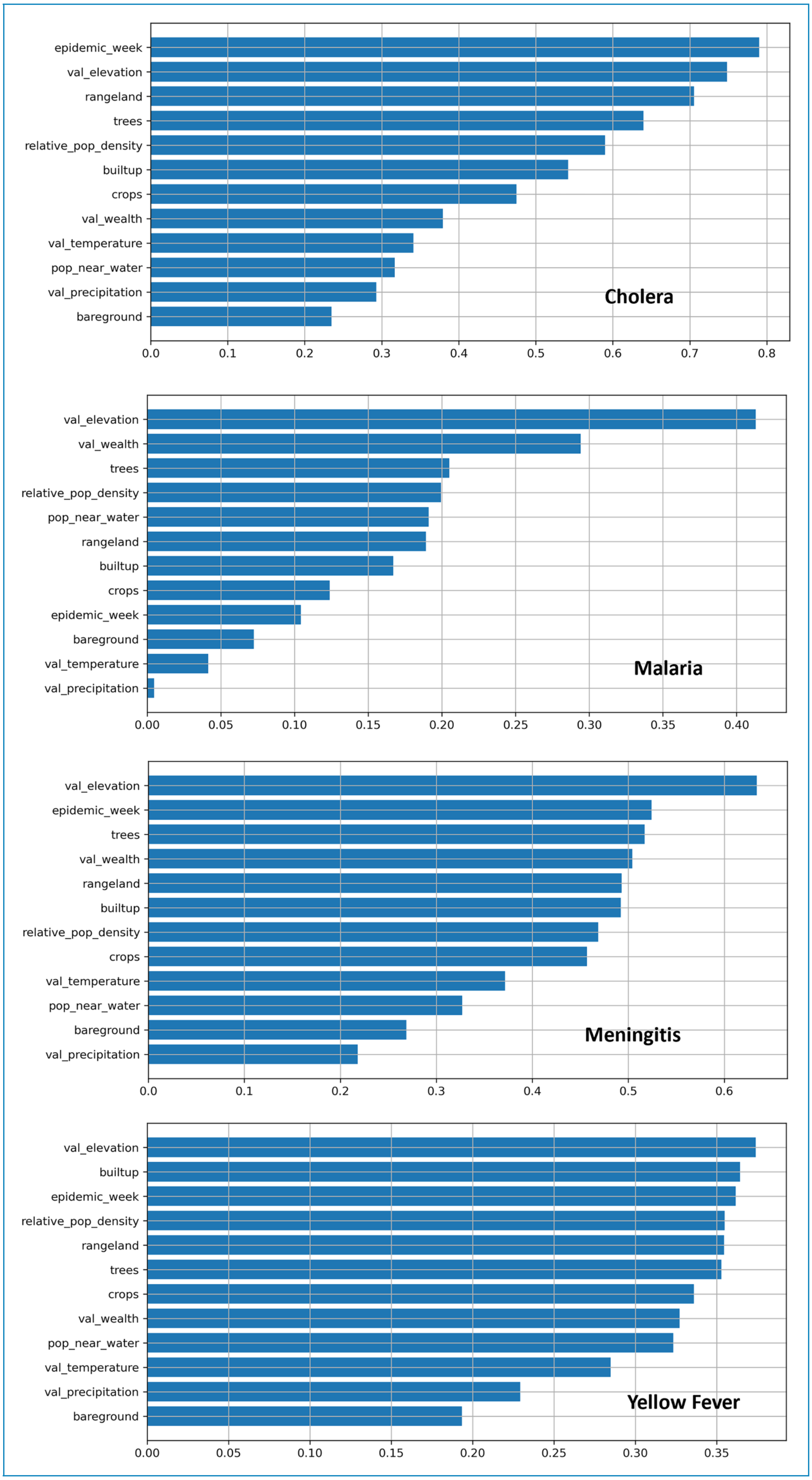

**Figure 3.** Results of machine learning feature importance measures of each disease. The values on the *X*-axis for feature importance are relative to each model; therefore, these values cannot be compared across disease charts.

machine learning will be vital in predicting and tackling disease spread.

### Intelligently selecting disease-affecting datasets

For discussion specific to the supporting datasets, the hot and cold spot analysis highlighted specific geographic areas more or less likely to influence disease case counts. Scientists can combine these results with feature importance measures to choose more supporting datasets to help machine learning model predictions.

Furthermore, a critical point to future data collection is that supporting datasets would best aid the machine learning model if they correspond by space and time to the IDSR dataset. Ideally, the supporting datasets should be at a spatial resolution fine enough to have a value for each district and a precise enough temporal resolution to have a value for each week. As noted in Table 3, our analysis over multiple years used only one data layer for elevation, two for population density, one for relative wealth, and four for land cover because the selected datasets are only available for these years. On a local level, these datasets likely stayed consistent during our study period. However, ideally, we plan to perform further analysis on more frequent data when that data becomes available. For example, a high-resolution population density dataset generated more frequently can be valuable for disease analysis considering seasonal and crisis event migrations.

### Proposed large geographic scale analysis

A final discussion topic is that we encountered two primary challenges because our analysis covered a vast geographic area. First, potentially infinite combinations of factors affect disease spread in all local areas in the study area, making data and analysis choices difficult. Second, datasets that cover the study area are naturally big, making many analyses computer resource-intensive and time-consuming.

To overcome these challenges, we propose future precise research using a large map scale (covering a relatively small ground area) analysis. The geographic area we mentioned in the Spatial proximity's impact on disease outbreaks subsection around the Great Lakes of Africa is a top candidate for such research. Focusing on a smaller area would mean shorter computer analysis wait times and a more in-depth understanding of data patterns. This analysis can also be done to ensure that it scales to larger areas given more time and resources. For example, our proposed research could incorporate datasets like roads and other transportation networks; length of commute and other movement data like modes of transportation; water bodies; rivers; detailed data about water bodies like water retention, drainage, and depth; population distribution; demographic data like age, education, sex, and income; and human lifestyle data like occupation, religion, mobility, and gatherings. Since model research, analysis, and development are time-consuming, focusing on a relatively small geographic area while ensuring computational methods scale would produce valuable and efficient models that still have the advantage of rapidly producing insights over vast areas.

We anticipate that such research can answer questions like: When clustering districts to find similar districts by data, do they have similar case counts? How do proximity and usage of transportation networks affect cases? What are the effects of human mobility levels and types of cases? Do water body characteristics like water retention length, depth, and salinity affect cases? How does population proximity to other spatial data like road networks and schools affect cases? How do residents' demographic data factors like age affect cases? How do human lifestyle factors affect cases?

## Conclusion

In conclusion, we showed the value of modern geospatial data analytics and machine learning techniques to understand the spatial distribution of disease outbreaks, uncover spatial patterns, predict future outbreaks and their intensities, and derive the most critical contributing factors to disease outbreaks.

Our work is an initial pilot study; future analyses can extend our results. We also laid the groundwork for suggested future analyses to improve disease outbreak prevention, preparedness, detection, monitoring, and response. Computational techniques to analyze and predict disease are critical for improved public health outcomes.

**Acknowledgements:** We appreciate the valuable suggestions of the WHO AFRO Emergency Preparedness and Response team and consultants throughout our research. In particular, we would like to acknowledge the helpful comments from Dr Godwin Ubong Akpan, Dr Okot Charles Lukoya, and Dr George Sie Williams.

We also want to acknowledge Dr Dani Arribas-Bel. He authored the free online short course "Geographic Data Science," including clear Python tutorials to perform ESDA.[36] We began our ESDA with one of his tutorials and followed many of its steps. Open courses and tutorials like this that propel science should be commended.

**Contributorship:** SP and EK researched the literature and conceived the study. All authors were involved in protocol development and data acquisition. SP and EK were involved in data analysis. SP wrote the first draft of the manuscript. All authors reviewed and edited the manuscript and approved the final version of the manuscript.

**Data availability statement:** To ensure the transparency and reproducibility of our research, a sample of the IDSR data, the source code to download and process the supporting datasets, and the source code to analyze the data and make predictions are all available at https://github.com/scottpez/disease-outbreak-predictions[49].

**Declaration of conflicting interests:** The author(s) declared no potential conflicts of interest with respect to the research, authorship and/or publication of this article.

**Ethical approval:** Our data is an aggregated indicator of the number of suspected cases and suspected deaths recorded by each district weekly. It contains no personally identifiable information, so consent is not required.

**Funding:** The author(s) received no financial support for the research, authorship, and/or publication of this article.

**Guarantor:** SP.

**ORCID iD:** Scott Pezanowski https://orcid.org/0000-0001-5470-4925